\documentclass[letterpaper, 10 pt, conference]{ieeeconf}  % Comment this line out if you need a4paper

\IEEEoverridecommandlockouts                              % This command is only needed if 
\usepackage{graphics} % for pdf, bitmapped graphics files
\usepackage{epsfig} % for postscript graphics files
\usepackage{mathptmx} % assumes new font selection scheme installed
\usepackage{times} % assumes new font selection scheme installed
\usepackage{amsmath} % assumes amsmath package installed
\usepackage{amssymb}  % assumes amsmath package installed
\usepackage{subfigure}
\usepackage{algorithm} %format of the algorithm
\usepackage{algorithmic} %format of the algorithm
\usepackage{multirow} %multirow for format of table
\usepackage{booktabs}
\usepackage{flushend}
\title{\LARGE \bf
Jointly Learning Agent and Lane Information for Multimodal Trajectory Prediction
}

\author{Jie Wang, Caili Guo*, Minan Guo and Jiujiu Chen% <-this % stops a space
\thanks{*Corresponding author: Caili Guo}
\thanks{J. Wang, C. Guo, M. Guo and J. Chen are with Beijing Laboratory of Advanced Information Networks, Beijing University of Posts and Telecommunications, Beijing, 100876, China. {\tt\small \{wj1e, guocaili, guominan, chenjiujiu\}@bupt.edu.cn}.}%
}

\begin{document}

\maketitle
\thispagestyle{empty}
\pagestyle{empty}

%%%%%%%%%%%%%%%%%%%%%%%%%%%%%%%%%%%%%%%%%%%%%%%%%%%%%%%%%%%%%%%%%%%%%%%%%%%%%%%%
\begin{abstract}
Predicting the plausible future trajectories of nearby agents is a core challenge for the safety of Autonomous Vehicles and it mainly depends on two external cues: the dynamic neighbor agents and static scene context. Recent approaches have made great progress in characterizing the two cues separately. However, they ignore the correlation between the two cues and most of them are difficult to achieve map-adaptive prediction. In this paper, we use lane as scene data and propose a staged network that \textit{Jointly learning Agent and Lane information for Multimodal Trajectory Prediction} (JAL-MTP). JAL-MTP use a \textit{Social to Lane} (S2L) module to jointly represent the static lane and the dynamic motion of the neighboring agents as instance-level lane, a \textit{Recurrent Lane Attention} (RLA) mechanism for utilizing the instance-level lanes to predict the map-adaptive future trajectories and two selectors to identify the typical and reasonable trajectories. The experiments conducted on the public Argoverse dataset demonstrate that JAL-MTP significantly outperforms the existing models in both quantitative and qualitative.

\end{abstract}

%%%%%%%%%%%%%%%%%%%%%%%%%%%%%%%%%%%%%%%%%%%%%%%%%%%%%%%%%%%%%%%%%%%%%%%%%%%%%%%%
\section{INTRODUCTION}

Self-Driving Vehicles (SDVs) are going to change the way we live by providing safe, reliable and effective transportation for everyone everywhere. Predicting the nearby agents’ trajectories is crucial for SDVs to understand the surrounding environment and make high-level decisions, which ensures the safety and comfort of SDVs \cite{paden2016survey}. The biggest challenge in trajectory prediction is that the future is inherently ambiguous. For predicting the target’s trajectory, we have to completely consider the target agent's intention, the current static surroundings and the interactions with nearby agents, to generate all plausible future trajectories based on the current scenario, namely multimodal trajectory prediction.

Although the existing multimodal trajectory prediction methods have shown competitive performance on the accuracy metrics, there are still two major deficiencies: 1) They consider the dynamic social interaction and the static scene information separately. However, the distribution of surrounding dynamic agents will simultaneously affect the drivable state of lanes, such as occupancy and congestion. Ignoring this can lead to unreasonable trajectory and even collision. 2) The patterns among the generated trajectories are not explicitly decoupled from the map, it will lead to mediocre map generalization capability. Typically, the regression-based \cite{liang2020learning, ye2021tpcn} and fixed anchor-based methods \cite{phan2020covernet, chai2019multipath} either concentrate on the main mode or restricted for a certain scene. The existing lane-based methods \cite{luo2020probabilistic, pan2020lane, khandelwal2020if, kim2021lapred} either consider a single lane or fuse all lanes. They do not explicitly take lanes as goals and are hard to generate map-adaptive trajectories. 

In this paper, we propose a novel network for multimodal trajectory prediction, namely JAL-MTP. Towards the two aforementioned issues, we use a proposed social to lane (S2L) module to jointly represent the dynamic agents and static lane context into the dynamic map consists of instance-level lanes. We intuitively take instance-level lanes as goals which ensures the goals is map-adaptive. Then we apply the proposed recurrent lane attention (RLA) mechanism to learn the features in instance-level lane and achieve lane-based predicting. Further, two selectors are used to score and identify the final trajectories. Finally, we evaluate the qualitative and quantitative performance of JAL-MTP on the public Argoverse dataset \cite{chang2019argoverse}.

Our contributions are summarized as follows,
\begin{itemize}
\item We propose a staged multimodal trajectory prediction network that first jointly represent the context cues as instance-level lanes, then model the evolution of the current motion on instance-level lanes to generate map-adaptive trajectories.
\item We propose a map representation method that fuse the information of dynamic nearby agents and static lane into a joint representation, which could reflect the map constraints well.
\item Our approach achieves SOTA quantitative performance for some categories of Argoverse benchmark and more reasonable multimodal qualitative results against the exist models.
\end{itemize}

\section{Related works}
In this section, we review works on modeling the two important cues for trajectory prediction: dynamic agents interaction and static scene context, conclude the existing methods that aim to multimodal trajectory prediction.
\subsection{Interaction among Dynamic Agents}
Related agents share the same scenario, co-operating with each other to perform safe actions. Accounting for the social interaction among neighboring agents is critical for safe and reasonable trajectory prediction. Grid-based methods \cite{li2019grip, deo2020trajectory} divide the space around agents into 2D grids and apply pooling to aggregate the interaction feature.These methods are hard to represent the irregular real-world interaction. Recently, graph-based methods \cite{mo2020recog} generalize deep learning on regular grids to arbitrary graphs with irregular topologies. They apply graph convolution \cite{jeon2020scale} or attention \cite{khandelwal2020if} to achieve message passing among the vertices composed of agents. These methods have achieved competitive performance in accuracy, however, they consider the social interactions independently, without utilizing the link between the agents and the scene.

\subsection{Static Scene Context Modeling}
Modeling the static scene context is the key to understand the intention and executable actions of the target agent. convolutional neural networks (CNN) based works \cite{phan2020covernet, chai2019multipath} use CNN to process the rasterized map image to obtain the scene context. These methods neglect the road topology such as the neighbor but opposite lanes. Recent methods use connective lanes sampled from the map as scene data. graph nerual networks (GNN) based methods\cite{liang2020learning, ye2021tpcn, gao2020vectornet} take the sampled points of lane centerline from the semantic map as vertices on a graph and obtain the scene context through GNN. They achieve competitive distance-based metrics but poor multimodal. lane-based methods\cite{luo2020probabilistic, kim2021lapred, khandelwal2020if} respectively propose a lane attention mechanism to use lane as scene data. MTPLA\cite{luo2020probabilistic} utilizes the most correlated lane. LaPred\cite{kim2021lapred} integrates all of the lane feature with learnable weights. WIMP\cite{khandelwal2020if} focus on the effective segments of the most correlated lane. We share a similar idea to lane-based methods. The difference is that we jointly represent each lane with their associated agents and separately treat each lane by proposed RLA.

\subsection{Multimodal Trajectory Prediction}
The future motion of agents is inherently multimodal, it requires to generate multiple trajectories with likelihoods. Some works are built upon stochastic models such as generative adversarial networks (GANs) \cite{sadeghian2019sophie} or variational autoencoders (VAEs) \cite{casas2020implicit}. Despite their competitive performance, the drawback of requiring multiple sampling to latent variables during inference prevents them from being applied. To avoid mode collapse \cite{makansi2019overcoming}, recent frameworks decompose the task into classification over anchor \cite{zhao2020tnt} or goal \cite{zhang2020map}, followed by conditional regression. TNT \cite{zhao2020tnt} take the lane centerline nodes as anchors which are accurate but too massive. GoalNet \cite{zhang2020map} propose multimodal generators to select goals among lanes but it mainly use rasterized map. Generally, lane information offers a strong prior on the semantic behavior of drivers. Driven by this, we take all possible lanes as goals and predict based on them.

\section{Approach}
In this section, we first introduce our problem formulation and notations in \ref{S31}. Then we present the details of JAL-MTP in \ref{S32}, which consists of three main modules what we called S2L, RLA Enc-Dec and selectors. Finally, we show our multimodal training strategy in \ref{S33}.
\subsection{Problem Formulation}\label{S31}
The purpose of trajectory prediction is to predict the future motion of all agents in a scene, given their past motion and current scene data. Since we focus on the multimodal problem, we only consider the one agent we called target in a scene. Suppose that the target agent has $N$ possible goals $G = \{ {g^i},i \in [1,N]\} $ on its situation, considering that drivers usually choose a potential goal $g^i$ to drive in reality and different goals can be treated independently, the task can be formulated as finding the posterior distribution of the future trajectory $Y$ given the possible goals $G$:
\begin{equation}
p({\rm{Y}}|G) = \sum\limits_{i = 1}^N {p({\rm{Y}}|{g^i})}\label{eq1},
\end{equation} 	
with approximating the distribution of ${\rm{Y}}$ as $K$ representative future trajectories ${\rm{Y}} = \left\{ {{Y^k},k \in [1,K]} \right\}$, where ${Y^k} = \{ y_t^k,t \in [1,{T_f}]\} $ denotes the $k$-th future trajectory over future time horizon ${T_f}$ and $y_{t}\in \mathbb{R}^{2}$ denotes the agent’s 2D coordinates in bird’s eye view(BEV) at time $t$.

The possible goals $G$ depend on the history motion state of the target agent, the distribution of nearby dynamic agents and static scene which are essentially drivable paths. The motion state of target agent and $M$ surrounding agents we called the social agents over the past ${T_P}$ timestamp can be formulated as ${\rm{X}} = \left\{ {{x_{t}}, t \in [ - {T_p} + 1,0]} \right\}$ and ${\rm{S}} = \left\{ {{S^j},j \in [1,M]} \right\}$ respectively, where ${S^j} = \left\{ {s_t^j, t \in [ - {T_p} + 1,0]} \right\}$ denotes the past trajectory of $j$-th social agent and $x_{t}, s_{t} \in \mathbb{R}^{2}$ denote the agents' BEV coordinates at time $t$. Similar to \cite{khandelwal2020if, kim2021lapred}, we describe the drivable paths as the concatenated centerline of lane segments, which called lane proposals corresponding to the $N$ goals and denoted as ${\rm{L}} = \left\{ {{L^i},i \in [1,N]} \right\}$, with ${L^i} = \left\{ {l_h^i,h \in [1,H]} \right\}$ and $l_{h}^{i} \in \mathbb{R}^{2}$ representing $h$-th lane node of $i$-th lane because the path are composed of discrete samples, $H$ is the length of a lane. 

As mentioned above, $G$ boll down to the sum of ${\rm{X}}$ , ${\rm{S}}$ and ${\rm{L}}$. In most cases the drivers tend to choose a lane proposal as their goal and drive along the centerline of the lane proposal. In addition, ${\rm{S}}$ and ${\rm{L}}$ are interrelated and mutually influenced, which can be specifically performed the distribution of surrounding dynamic agents will simultaneously affect the drivability of lane proposals. Therefore, we comprehensively aggregate the social agents’ situation ${\rm{S}}$ into lane proposals ${\rm{L}}$ to obtain instance-level lanes ${\rm{\tilde L}} = \left\{ {{{\tilde L}^i},i \in [1,N]} \right\}$, then we can regard the $N$ goals as the evolution results of the current motion state ${\rm{X}}$ on $N$ instance-level lanes, so \eqref{eq1} could be expressed as
\begin{equation}
p({\rm{Y}}|G) = \sum\limits_{i = 1}^N {p({\rm{Y}}|{\rm{X}},{\rm{S}},{L^i})}  = \sum\limits_{i = 1}^N {p({\rm{Y}}|{\rm{X}},{{\tilde L}^i})}\label{eq2},
\end{equation} 	
where ${\tilde L^i}$ represent $i$-th instance-level lane under ${g^i}$, we can adequately characterize the distribution of future trajectories by aggregating the trajectory predicted under $N$ instance-level lanes. We consider the multimodal trajectory prediction problem and build our model driven by Eq.~\eqref{eq2}.

\subsection{Model Structure}\label{S32}
\begin{figure*}[ht] 
\includegraphics[width=1\textwidth]{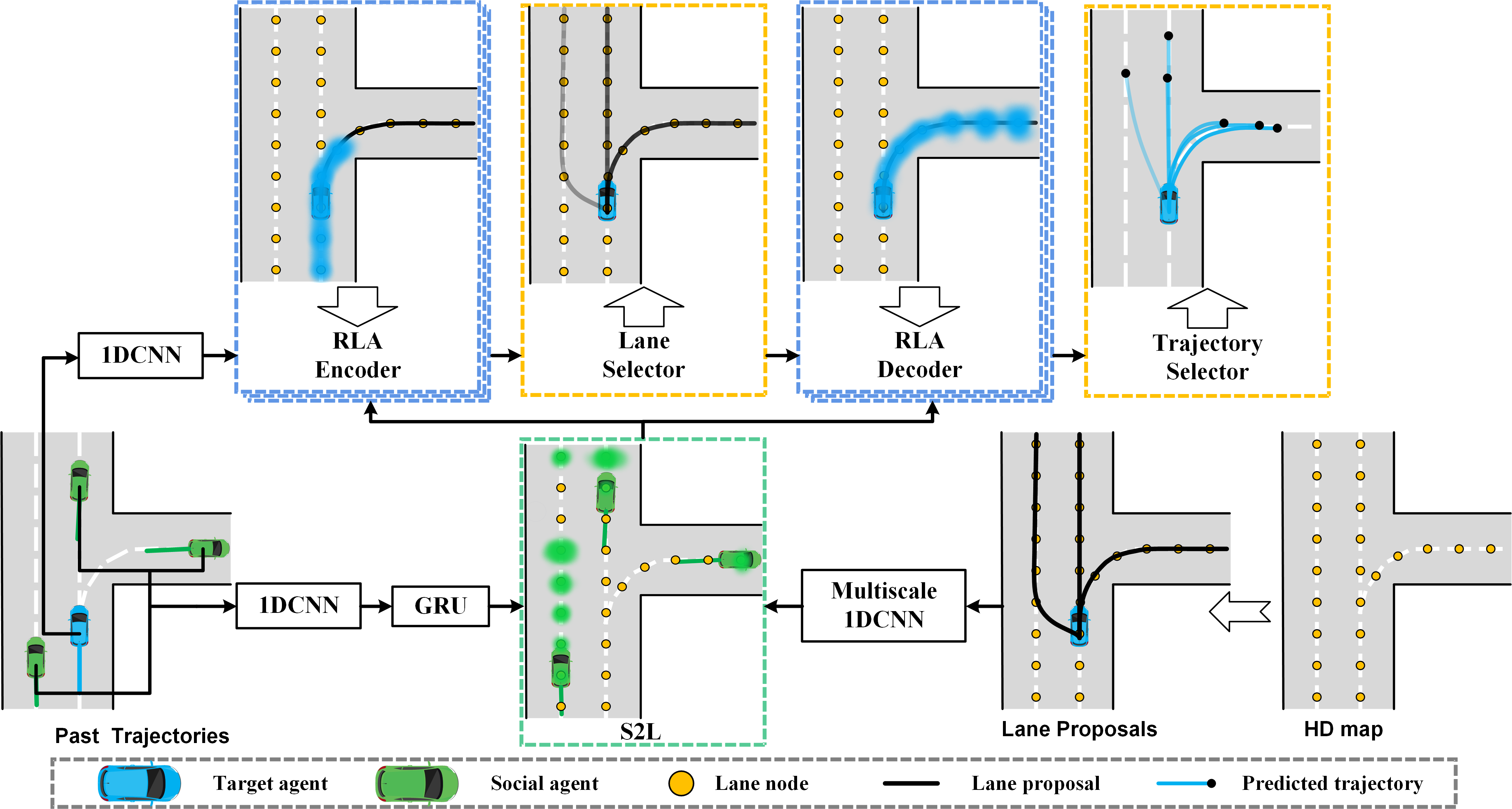}
\caption{Overall architecture: The model consists of three major modules, RLA encoder-decoder(Enc-Dec), S2L and two selectors (respectively rendered as blue, green and yellow). The inputs to the network are set of target and social agents’ past trajectories (respectively marked as blue and green lines) and sampled lane nodes (marked as yellow circles). For the sampled lane nodes, we employ a heuristic lane proposal algorithm based on the map preprocess methods proposed in \cite{chang2019argoverse} to find the lane proposals.}
\label{fig1}
\end{figure*}   
%\subsubsection{Overall Architecture}\label{S321}
The model structure of the proposed network is corresponding to the formulation expressed in \ref{S31}. Fig.~\ref{fig1} depicts the overall structure of our network. The lane proposals obtaind by preprocessing and the social agents’ past trajectories are respectively fed into the multi-scale one-dimensional Convolutional Neural Network (1D-CNN) and the 1D-CNN followed by Gated Recurrent Unit (GRU) to encode as the lane proposals feature ${\rm{L}}$ and social motion feature ${\rm{S}}$. For effectively representing the road information as instance-level lane ${\tilde {\rm{L}}}$, we apply a spatial attention in S2L to jointly fusing the ${\rm{L}}$ and  ${\rm{S}}$. The instance-level lane will serve as the extra input to RLA modules. For the target agent, by applying the 1DCNN followed by RLA Encoder, we can encode the past motion feature ${\rm{X}}$ and current lane condition ${\tilde L^i}$ to hidden vector $h^i$. By feeding the vector to Lane Selector, we can get the lane score which marked as the transparency. Finally, based on the hidden vector and lane score, the RLA Decoder and Trajectory Selector will produce and select K typical trajectories with probabilities.
\begin{figure}[htbp]
\includegraphics[width=0.5\textwidth]{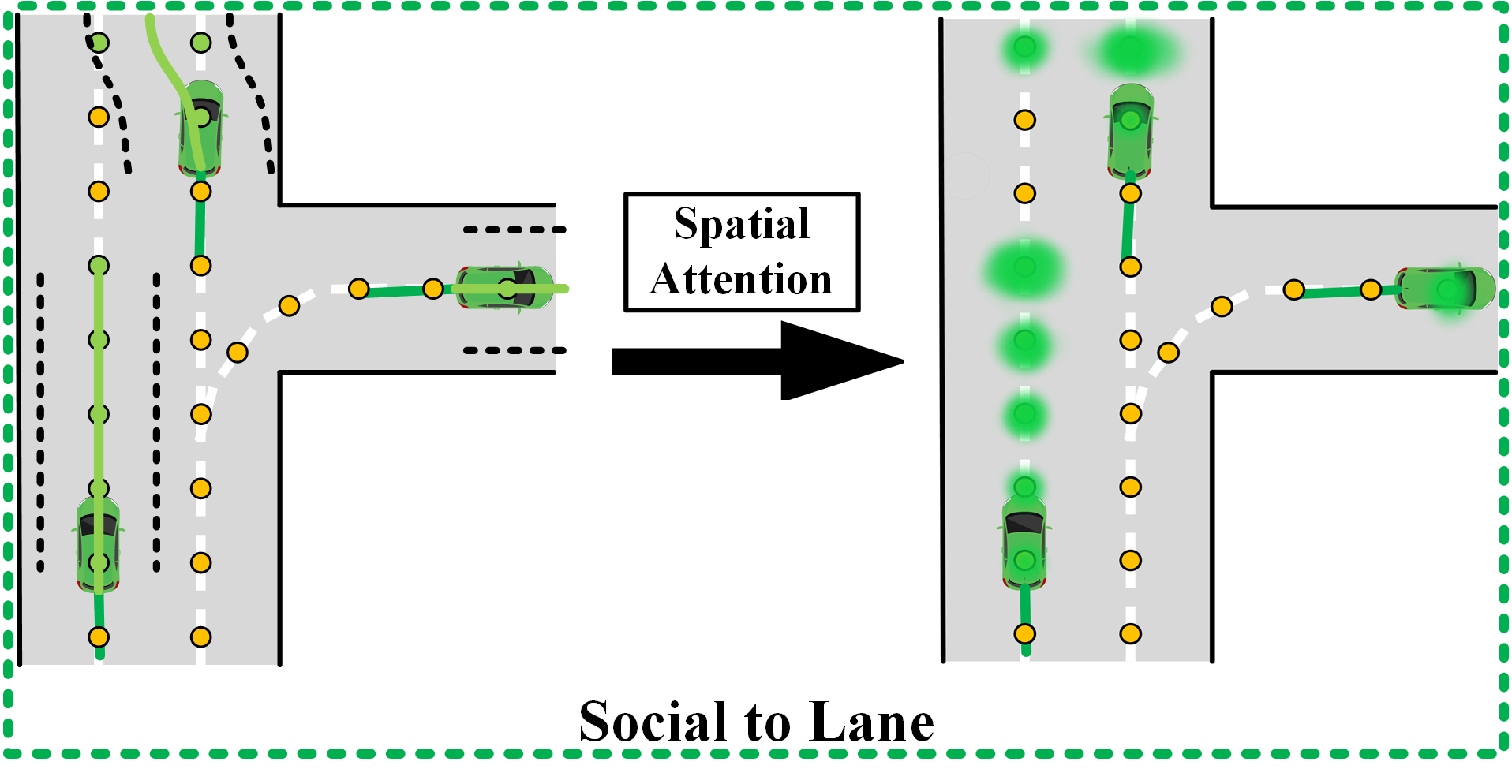}
\caption{The process of Social to Lane module. Light green line denotes the trajectory rollout of social agents. The related lane nodes are stained green.}
\label{figS2L}
\end{figure} 
\subsubsection{S2L: Jointly representing the agent and lane information }\label{S322}
The S2L devotes to fusing the ${\rm{S}}$ and ${\rm{L}}$ to extract the joint agents-lane representation ${\tilde {\rm{L}}}$ by the spatial attention mechanism we proposed. The process is shown in Fig.~\ref{figS2L}. Firstly, we apply the constant acceleration kinematic equation to simply calculate the social agents’ future trajectory rollout from their current motion states. Given the trajectory rollouts, we can find the related lane nodes to apply the spatial attention:
\begin{equation}
\hat{l}_{h}^{i}=\varphi_{r e s}\left(l_{h}^{i}+\sum_{j=1}^{M} \varphi_{a g g}\left(l_{h}^{i}\|\operatorname{dist}\| S^{j}\right)\right),\label{eq3}
\end{equation} 	
where $h \in [1,H]$, $l_h^i$ and $\hat l_h^i$ respectively denote $h$-th lane nodes of the $i$-th lane proposal and instance-level lane, ${S^j}$ denotes the feature of the related social agent whose trajectory rollout’s ${l_2}$ distance from the lane nodes is smaller than a threshold (e.g. 7.5m), ${\varphi _{res}}( \cdot )$ and ${\varphi _{agg}}( \cdot )$ are  multi-layer perceptron(MLP) whose weights are shared over all nodes, $\parallel $ is concatenation, and $dist$ represent the spatial relationship between the lane nodes and related social agent, where we use $MLP({V_{l_{h}^i}} - {V_{{S^j}}})$ with $V$ representing the coordinates.
\begin{figure}[htbp]
\includegraphics[width=0.5\textwidth]{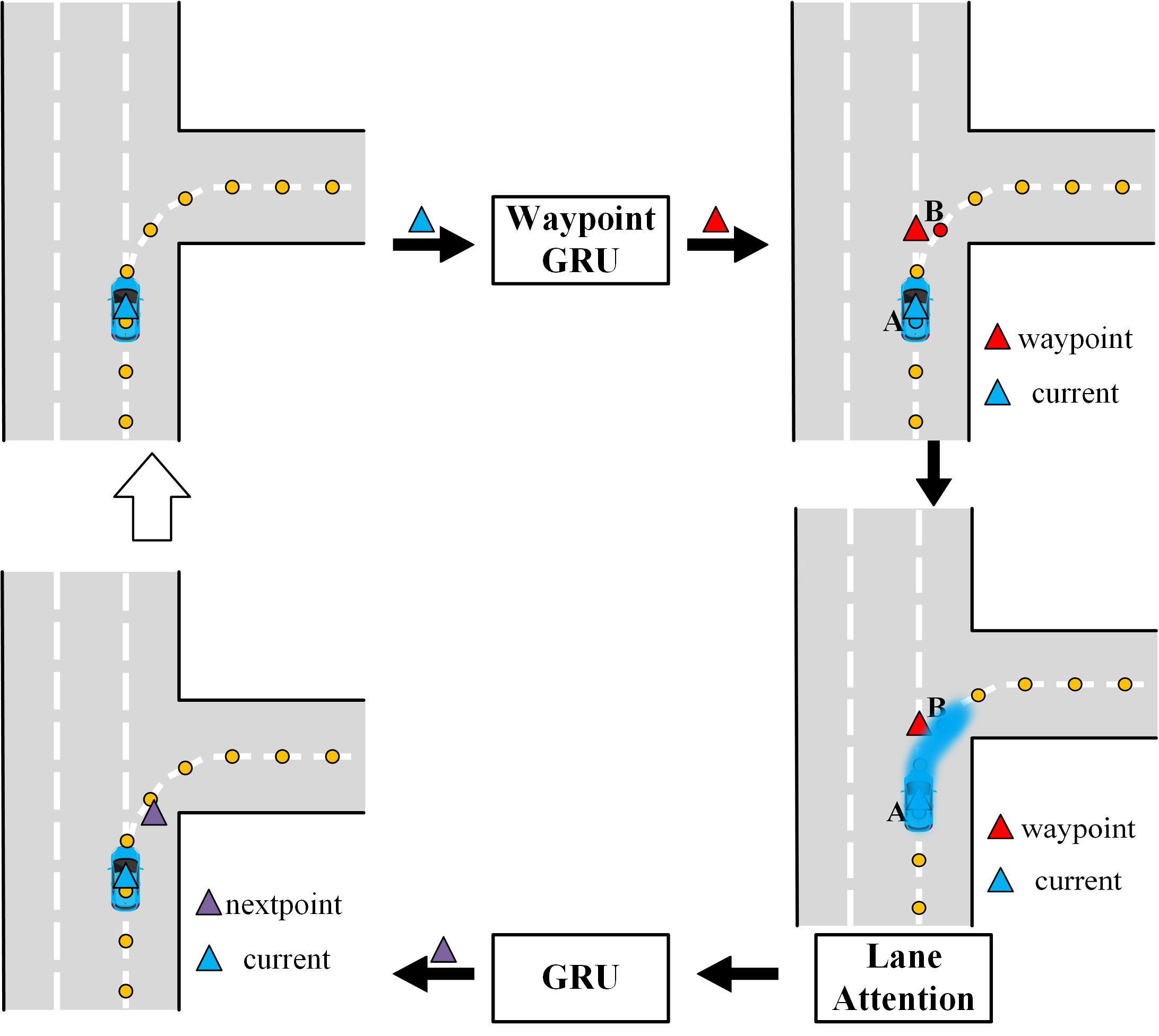}
\caption{The RLA Enc-Dec Module include three blocks: waypoint GRU, Lane Attention, GRU. This lane attention process is similar to \cite{khandelwal2020if}, but we additionally consider the relative position information. Note that both the waypoint GRU and the GRU have two layers.}
\label{fig3}
\end{figure} 	

After aggregate the related social feature to lane nodes by Eq.~\eqref{eq3}, we pass the updated lane nodes into another 1D-CNN to perform message passing among the lane nodes like GNN and get the final updated instance-level lane $\tilde l_{h}^i$. Then the produced $\tilde l_{h}^i$ will be used by RLA to help neural network understand the current scene context.
\subsubsection{RLA Enc-Dec: Learning from the instance-level lane representation}\label{S323}
Taking the motion feature $\rm{X}$ from 1DCNN and instance-level lane $\tilde l_{h}^i$ from S2L as input, the RLA module aim to generate future trajectories based on taking the input lane as goal. Intuitively, drivers usually pay more attention to the area they currently close to and the area to be reached. As shown in Fig.~\ref{fig3}, firstly, waypoint GRU takes the target agent’s motion history ${h_{t - 1}}$ together with the current location ${a_{t - 1}}$ (blue  triangle) as input to update the hidden state ${h_t}$ and predict the waypoint ${b_t}$ (red  triangle) according to
\begin{equation}
{b_t} = {\varphi _{{\rm{waypoint}}}}({h_t}){\rm{  }},{\rm{  }}{h_t} = GR{U_{{\rm{waypoint}}}}({a_{t - 1}},{h_{t - 1}})\label{eq4},
\end{equation} 	
where ${\varphi _{waypoint}}( \cdot )$ is MLP.
Based on the ${a_{t - 1}}$ and the ${b_t}$, we can find the nearest Euclidean points A (marked as blue) and B (marked as red) on $i$-th lane. Then we can perform goal-based prediction conditioned on the interested lane nodes from A to B via lane attention:
\begin{equation}
\begin{array}{l}
{Q^i} = h_{t - 1}^i{W^Q},\\
K_{A:B}^i = E_{A:B}^i{W^K},\\
V_{A:B}^i = E_{A:B}^i{W^V},
\end{array}\label{eq5}
\end{equation} 	
where ${W^Q}$, ${W^K}$, ${W^V}$ are learned weights, ${Q^i}$ relates to the motion history and the $K_{A:B}^i$ and $V_{A:B}^i$ are calculated from the states of the lane nodes A to B together with their spatial information relative to the agent’s current motion, which can be illustrated as follows:
\begin{equation}
\begin{array}{l}
R_{A:B}^i = MLP(concat(D_{A:B}^i,\Delta \theta _{A:B}^i)),\\
E_{A:B}^i = MLP(concat(\tilde l_{A:B}^i,R_{A:B}^i)),{\rm{ }}
\end{array}\label{eq6}
\end{equation} 	

where $D_{A:B}^i$ is the Euclidean distance between the ${a_{t - 1}}$ and the related lane nodes $l_{A:B}^i$, $\Delta \theta _{A:B}^i$ is the angle difference between the agent’s current orientation and the curvature of lane segment A to B, so $R_{A:B}^i$ represents the relative spatial information changes along A to B and $E_{A:B}^i$ covers the joint interaction features and spatial variation of A to B on lane $i$ , which corresponds to the goal-based intuition. Then the updated hidden states can be obtained by adding the attention results ${\alpha ^i}$:
\begin{equation}
\begin{array}{l}
{\alpha ^i} = \sum\limits_{j = A}^B {softmax(\frac{{{Q^i}K{{_j^i}^T}}}{{\sqrt {{d_k}} }})V_j^i}, \\
\tilde h_{t - 1}^i = h_{t - 1}^i + {\alpha ^i},
\end{array}\label{eq7}
\end{equation} 	

Finally, the updated hidden states are entered into the GRU followed by the predictor to predict the next trajectory point $\hat a_t^i$ (purple  triangle):
\begin{equation}
\hat a_t^i = {\varphi _{predictor}}(h_t^i){\rm{  }},{\rm{  }}h_t^i = GR{U_{{\rm{next}}}}(a_{t - 1}^i,\tilde h_{t - 1}^i),\label{eq8}
\end{equation} 	
where ${\varphi _{predictor}}( \cdot )$ is MLP. Note that for the RLA encoder,  $t=[- {T_p} + 1,0]$  to learn how to encode the motion information and scene states over the past time into the hidden vector, but for the RLA decoder, $t=[1,{T_f}]$  to learn how to adjust to the changes of the real time scene to make multimodal and map-adaptive prediction. 

\subsubsection{Selectors: Identifying typical trajectories}\label{S324}
\begin{figure}[htbp]
\includegraphics[width=0.5\textwidth]{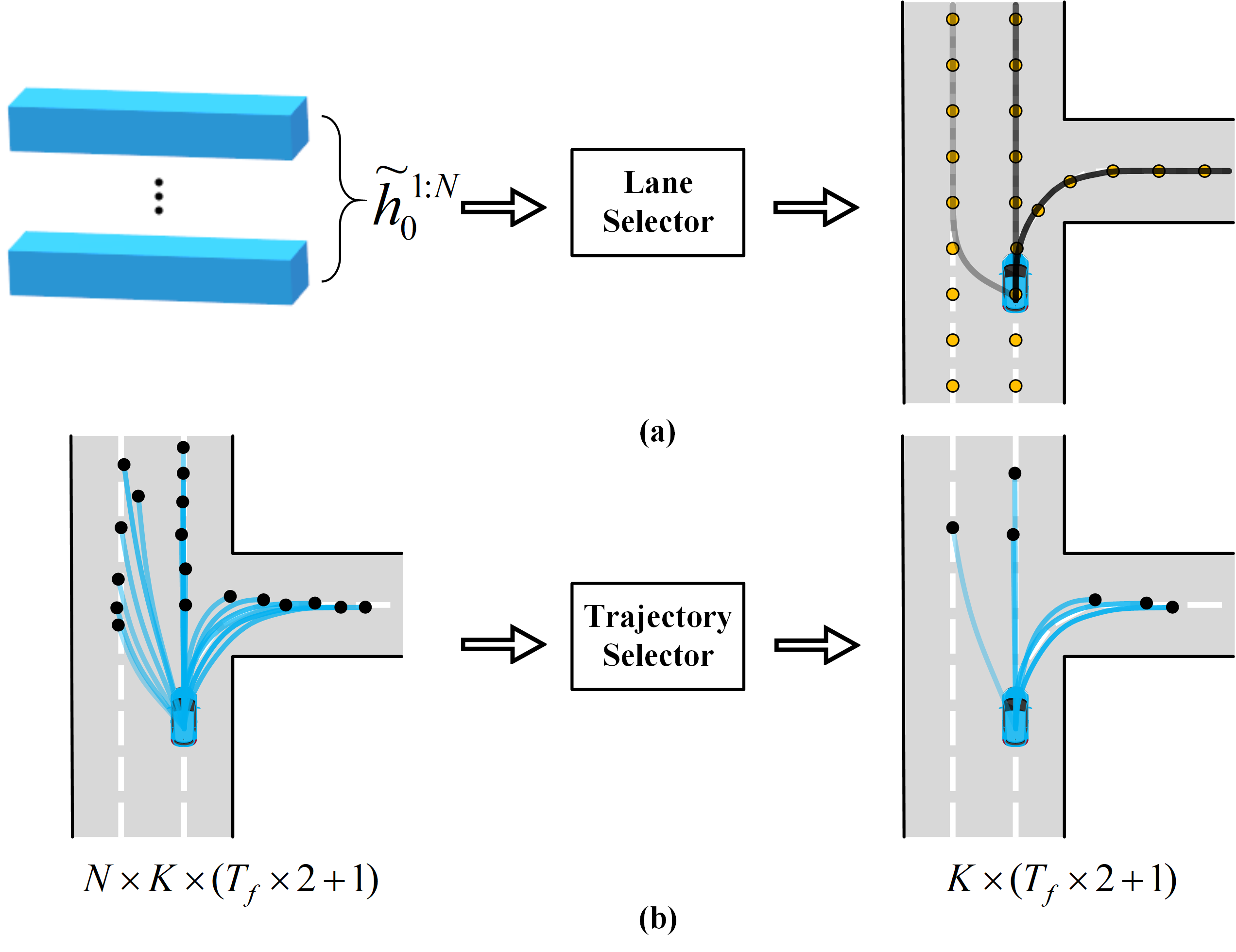}
\caption{Selectors for lane (a) and trajectory (b).}
\label{fig4}
\end{figure}

As shown in Fig.~\ref{fig4}, the lane selector takes the $N$ hidden vector from the RLA encoder as input to obtain $N$ scores of $N$ lane. The trajectory selector scores the $N \times K$ trajectories from the RLA decoder and multiply the trajectory scores by the corresponding lane scores to get the final $N \times K$ trajectory scores. Finally, we intercept the $K$ highest score trajectories as the final trajectories to calculate the metrics for evaluation, and normalize the $K$ trajectory scores as the corresponding probability. 
Note that both the lane selector and the trajectory selector are two-layer MLP.

\subsection{Training}\label{S33}
The regression and two classification tasks are independent and the whole model can be trained end-to-end with a total loss function: 
\begin{equation}
{L_{total}} = {L_{reg}} + {\lambda _1}{L_{lanecls}} + {\lambda _2}{L_{trajcls}},\label{eq9}
\end{equation} 	
where ${\lambda _1}$ and ${\lambda _2}$ are hyperparameters to balance the tasks.

For regression, we apply the ${l_1}$ loss on all predicted time steps that proved effective in previous works \cite{khandelwal2020if}:
\begin{equation}
{L_{reg}} = \frac{1}{{{T_f}}}\mathop {\min }\limits_{k \in \{ 1...K\} } \sum\limits_{t = 1}^{{T_f}} {\left\| {{{\rm{y}}_t} - {\rm{\hat y}}_t^k} \right\|},\label{eq10}
\end{equation} 
where $\hat y_t^k$ denotes the future position of the k-th trajectory at time t and ${y_t}$ is the corresponding ground truth.

For classification, we use the cross-entropy loss between the output scores and the labels: 
\begin{equation}
\begin{array}{l}
{L_{lanecls}} = {L_{CE}}(scor{e_{lane}},labe{l_{lane}}),\\
{\rm{ }}{L_{trajcls}} = {L_{CE}}(scor{e_{traj}},labe{l_{traj}}),
\end{array}\label{eq11}
\end{equation} 
where the labels are generated during training with self-supervised learning task by
\begin{equation}
\begin{array}{l}
labe{l_{lanecls}} = \frac{{\exp ( - {D_1}(L,{{\rm{Y}}_{GT}}))}}{{\sum\limits_{i = 1}^N {\exp ( - {D_1}({L_i},{{\rm{Y}}_{GT}}))} }} ,\\
labe{l_{trajcls}} = \frac{{\exp ( - {D_2}({\rm{Y}},{{\rm{Y}}_{GT}}))}}{{\sum\limits_{i = 1}^K {\exp ( - {D_2}({{\rm{Y}}_i},{{\rm{Y}}_{GT}}))} }} ,
\end{array}\label{eq12}
\end{equation}
with $D_{1}\left(L, Y_{G T}\right)=\sum_{t=1}^{T_{f}} \beta(t) \min _{h \in\{1, \ldots, H\}}\left\|y_{t}-l_{h}\right\|$ and $D_{2}\left(\mathrm{Y}, \mathrm{Y}_{G T}\right)=\sum_{t=1}^{T_{f}}\left\|y_{t}-\hat{y}_{t}\right\|$ denoting the distance between the ground truth trajectory and the lane, the prediction respectively, where $\beta (t) = t$ is the scaling weight to indicate the importance of different time steps. Note that there is only one ground truth trajectory, in order to make the model better learn to follow a certain lane, we only predict the trajectory under the most likely lane during training, when inference, we input all of the lane proposal and apply the selectors to get final trajectories.

\section{Experiments}
In this section, we evaluate the quantitative and qualitative performance of the proposed model on the Argoverse motion forecasting benchmark. 
\subsection{Experimental Setup}\label{S41}
\subsubsection{Dataset}\label{S411}
Argoverse is a motion forecasting dataset with 325K 5 second trajectory sequences collected in Pittsburgh and Miami. The sequences are split into training, validation and test sets, which have 206K, 39K and 78K sequences respectively. The trajectories comprising of 2D coordinate sequences are sampled at 10Hz. Each sequence has one interesting object called “agent”, the task is using the agents' past 2 seconds trajectory data to predict the future locations of agents in 3 seconds. In addition to trajectories, each sequence is associated with a high definition map composed of lane centerline sets and their connectivity. Importantly, Argoverse dataset is collected for interesting and diverse behaviors, which is suitable for the study of multimodal trajectory prediction. Because of the test set without labels, we conduct our experiments mainly on the validation set. 
\subsubsection{Metrics}\label{S412}
To evaluate a multimodal set of predicted trajectories, we take the best trajectory among the $K$ generated trajectories to calculate the minimum average displacement error ($\operatorname{minADE}_{K}$) and final displacement error ($\operatorname{minFDE}_{K}$) : 
\begin{equation}
\begin{array}{l}
\min \mathrm{ADE}_{K}=\frac{1}{T_{f}} \mathop {\min _{k \in\{1, \ldots, K\}}} {\sum\nolimits_{t = 1}^{{T_f}} {\left\| {{y_t} - \hat y_t^k} \right\|} _2} ,\\
\min {\rm{FD}}{{\rm{E}}_K} = \mathop {\min _{k \in\{1, \ldots, K\}}} {\left\| {{y_{{T_f}}} - \hat y_{{T_f}}^k} \right\|_2} ,
\end{array}\label{eq10}
\end{equation} 
where $\hat y_t^k$ denotes the future position of the k-th trajectory at time t and ${y_t}$ is the corresponding ground truth. We also report miss rate (MR) to measure the ratio of scenarios where none of the predictions are within 2 meters of the ground truth according to $\operatorname{minFDE}_{K}$. These metrics indicate the accuracy of the predicted trajectories. To evaluate the performance of our selectors, we also calculate  the probabilistic-based metrics (p-FDE, p-ADE, p-MR, brier-FDE, brier-ADE) [10], which measures the reasonability of predicted probabilities and not be elaborated because of only used in ablation study. Notably, for all metrics, the smaller is the better.

\subsection{Quantitative Results}\label{S42}

We compare JAL-MTP with the existing methods that have similar configuration and SOTA performance. Table.~\ref{table:Result} presents the results on the validation set with $K = 1,6$. GRU ED is the baseline that simply GRU encoder-decoder structure. MTPLA \cite{luo2020probabilistic}, WIMP \cite{khandelwal2020if} and LaPred \cite{kim2021lapred} are lane-based methods that also use instance-level lane similar to us, the difference is that MTPLA and LaPred map the entire lane to a vector, WIMP consider without the explicit changes in position and curvature of the lane segment. LaneGCN \cite{liang2020learning} and TPCN \cite{ye2021tpcn} are the GNN-based models that have better performance in accurate metrics through GNN but perform bad in multimodal results. The quantitative results show that our method out-performs all related lane-based works in almost all metrics and achieves similar performance to the GNN-based models. 
Notably, our ${\rm{minFD}}{{\rm{E}}_6}$ and MR metric are the best among all baselines, which indicates that our method has a competitive ability to capture the main mode while considering the multimodal of predicted trajectories.
\begin{table}[htbp]
\caption{Results on Argoverse motion forecasting validation set.} % title of Table
\centering % used for centering table
\setlength{\tabcolsep}{0.9mm}{
\begin{tabular}{c|c c c c c} % centered columns (4 columns)
\hline\hline %inserts double horizontal lines
Model & ${\rm{minFD}}{{\rm{E}}_1} $ & ${\rm{minAD}}{{\rm{E}}_1}$ & ${\rm{minFD}}{{\rm{E}}_6}$ & ${\rm{minAD}}{{\rm{E}}_6}$ & $\operatorname{MR}$ \\ [0.5ex] % inserts table
%heading
\hline % inserts single horizontal line
GRU ED & 3.75 & 1.64 & 1.71 & 0.97 & 0.23 \\ % inserting body of the table
\midrule[0.5pt]
MTPLA & 3.27 & 1.46 & 2.06 & 1.05 & -  \\
LaPred & 3.29 & 1.48 & 1.44 & \textbf{0.71} & - \\
WIMP & 3.19 & 1.45 & 1.14 & 0.75 & 0.12 \\
\midrule[0.5pt]
TPCN & \textbf{2.95} & \textbf{1.34} & 1.15 & 0.73* & 0.11* \\ 
LaneGCN & 2.97* & 1.35* & 1.08* & \textbf{0.71} & 0.11* \\ 
\midrule[1pt]
Ours & 3.00 & 1.39 & \textbf{1.07} & 0.73* & \textbf{0.10} \\ [1ex]% [1ex] adds vertical space
\hline %inserts single line
\end{tabular}}
\label{table:Result} % is used to refer this table in the text
\end{table} 
\begin{table*}[ht]
\setlength\tabcolsep{4.7pt}
\caption{Ablation study results of modules} % title of Table
\centering % used for centering table
\begin{tabular}{c c c c|c c c c c c c c} % centered columns (4 columns)
\hline\hline %inserts double horizontal lines
% & & Modules & & & & & K=6 & & & \\ [0.5ex] % inserts table
\multicolumn{4}{c|}{Modules} & \multicolumn{8}{c}{K=6} \\ 
%heading 
\cline{1-4} % inserts single horizontal line
LA & Position & S2L & Selectors& $\operatorname{minFDE}_{K}$ & $\operatorname{minADE}_{K}$ & MR & p-FDE & p-ADE & p-MR & brier-FDE & brier-ADE \\ \hline % inserting body of the table 
 & & & & 1.71 & 0.97 & 0.233 & 3.50 & 2.76 & 0.872 & 2.40 & 1.66 \\ \hline
$\surd$ & & & & 1.27$_{\downarrow 0.44}$ & 0.80$_{\downarrow 0.17}$ & 0.138$_{\downarrow 0.095}$ & 3.06$_{\downarrow 0.44}$ & 2.59$_{\downarrow 0.17}$ & 0.856$_{\downarrow 0.016}$ & 1.96$_{\downarrow 0.44}$ & 1.49$_{\downarrow 0.17}$ \\ \hline
$\surd$ & $\surd$ & &  & 1.20$_{\downarrow 0.07}$ & 0.77$_{\downarrow 0.03}$ & 0.122$_{\downarrow 0.016}$ & 2.99$_{\downarrow 0.07}$ & 2.56$_{\downarrow 0.03}$ & 0.853$_{\downarrow 0.003}$ & 1.89$_{\downarrow 0.07}$ & 1.47$_{\downarrow 0.02}$ \\ \hline
$\surd$ & $\surd$ & $\surd$ & & \textbf{1.07}$_{\downarrow 0.13}$ & \textbf{0.73}$_{\downarrow 0.04}$ & \textbf{0.098}$_{\downarrow 0.024}$ & {2.87}$_{\downarrow 0.12}$ & {2.53}$_{\downarrow 0.03}$ & {0.849}$_{\downarrow 0.004}$ & {1.77}$_{\downarrow 0.12}$ & {1.43}$_{\downarrow 0.04}$\\ \hline
$\surd$ & $\surd$ & $\surd$ & $\surd$ & \textbf{1.07}$_{\downarrow 0.00}$ & \textbf{0.73}$_{\downarrow 0.00}$ & \textbf{0.098}$_{\downarrow 0.000}$ & \textbf{2.73}$_{\downarrow 0.14}$ & \textbf{2.38}$_{\downarrow 0.15}$ & \textbf{0.817}$_{\downarrow 0.032}$ & \textbf{1.71}$_{\downarrow 0.06}$ & \textbf{1.37}$_{\downarrow 0.06}$\\ [1ex] % [1ex] adds vertical space
\hline %inserts single line
\end{tabular}
\label{table:Ablation} % is used to refer this table in the text
\end{table*}
\begin{figure*}[ht] 
\includegraphics[width=1\textwidth]{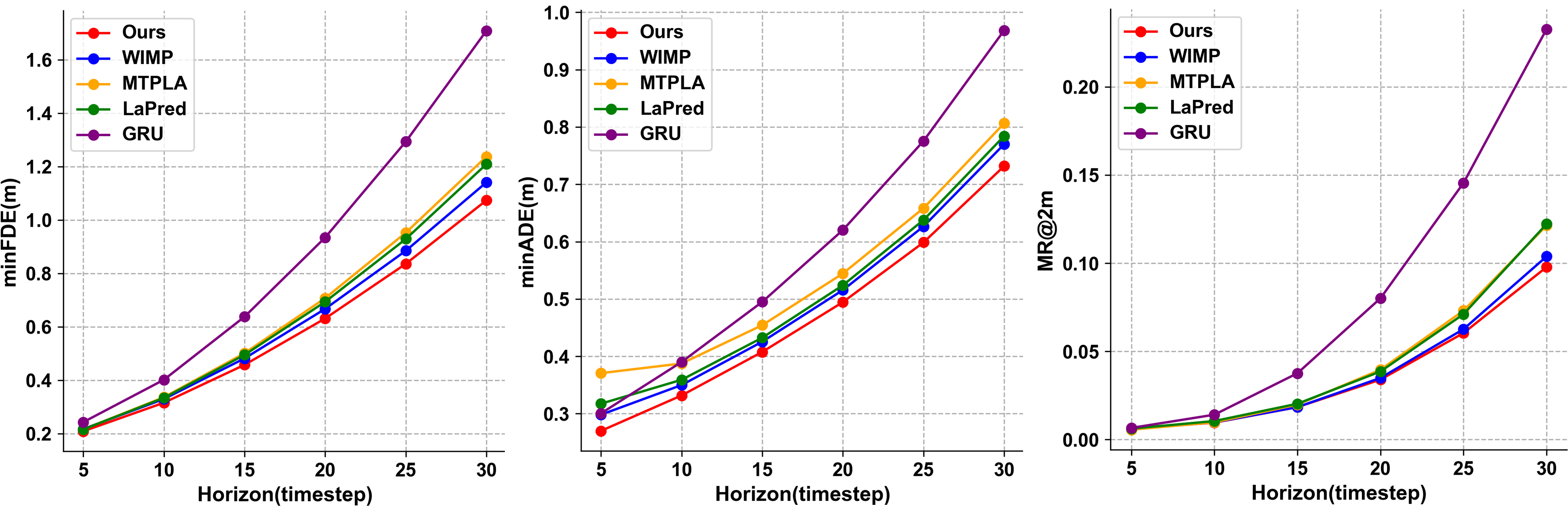}
\caption{A comparison of metrics at different forecasting horizons. All the models are introduced in \ref{S42} and all of them except GRU are lane-based.}
\label{figtime}
\end{figure*}

\subsection{Ablation study}\label{S43}
\subsubsection{Importance of each module}\label{S431}
In Table.~\ref{table:Ablation}, we show the results of using GRU encoder-decoder as baseline (first line) and progressively adding the rest of modules to the network.

The RLA module is divided into LA and Position. The Selectors column denotes using the two selectors to mark and identify typical trajectories rather than treat them equally. From results, all modules have contribution to the performance improvement. It illustrates the efficiency of our modules to predict accurate trajectories. Moreover, the inclusion of LA is more effective for boosting the performance. It means the lane information is more critical for prediction. Notably, the improvement in probabilistic-based metrics by Selectors proves its ability to select reasonable trajectories.

\subsubsection{Superiority of our methods}\label{S432}
For fair contrast, We only change the different lane attention or social interactive extraction methods based on our whole model in Table.~\ref{table:Superiority} Lane and Social section. In Lane section, LA-MTPLA and LA-LaPred respectively map the whole lane nodes sequence into a hidden vector by pooling and LSTM. LA-overall apply attention to all lane nodes. LA-WIMP execute attention on the local lane nodes similar to us, but without the Position information. \begin{table}[htbp]
\caption{Comparative study on lane and social \\
modeling methods} % title of Table
\centering % used for centering table
\setlength{\tabcolsep}{1.5mm}{
\begin{tabular}{c|c|c c c} % centered columns (4 columns)
\hline\hline %inserts double horizontal lines
 & Methods & ${\rm{minFD}}{{\rm{E}}_6} $ & ${\rm{minAD}}{{\rm{E}}_6} $ & MR \\ % inserts table
%heading
\hline % inserts single horizontal line
\multicolumn{1}{c|}{\multirow{5}{*}{Lane}} & LA-MTPLA & 1.24 & 0.81 & 0.121 \\\cline{2-5}
 & LA-LaPred & 1.21 & 0.78 & 0.122 \\\cline{2-5}
  & LA-overall & 1.15 & 0.75 & 0.112 \\\cline{2-5}
  & LA-WIMP & 1.12 & 0.75 & 0.104 \\\cline{2-5}
  & RLA & {\textbf{1.07}} & {\textbf{0.73}} & {\textbf{0.098}} \\ \hline 
 \multicolumn{1}{c|}{\multirow{2}{*}{Social}} & Separately by GAT & 1.20 & 0.78 & 0.123 \\ \cline{2-5}
   & Jointly by S2L & {\textbf{1.07}} & {\textbf{0.73}} & {\textbf{0.098}}\\[1ex] % [1ex] adds vertical space
\hline %inserts single line
\end{tabular}}
\label{table:Superiority} % is used to refer this table in the text
\end{table}
By comparing LA-overall, LA-WIMP, RLA with LA-MTPLA, LA-LaPred, it can be concluded that regarding the lane as a set of lane nodes is more effective than simply mapping it to a vector. The best performance achieved by RLA reveals the superiority of our lane attention method. In Social section, Separately by GAT means treating the agents as vertices on a fully-connected graph and performing attention to independently model the interaction. The significant improvement on all metrics by S2L proves the effective of our S2L in jointly representing the social and lane information. 

\subsubsection{Impact of forecast horizon}\label{S433}
In Fig.~\ref{figtime}, our model outperforms all baselines across all horizons. Furthermore, the gap between our model and baselines gradually widens as the horizon increasing, which underscores the ability to longer-term prediction of our model. GRU baseline achieves a very low minADE in time step 5, but quickly grow beyond the lane-based methods including us in longer horizon. It reveals the short-term behavior of an agent is easier to model with fewer constrains and simpler models, but the long-term behavior needs deeper attributes such as its goal to achieve.
\subsection{Qualitative analysis}\label{S44}
\begin{figure*}[htbp] 
\includegraphics[width=1\textwidth]{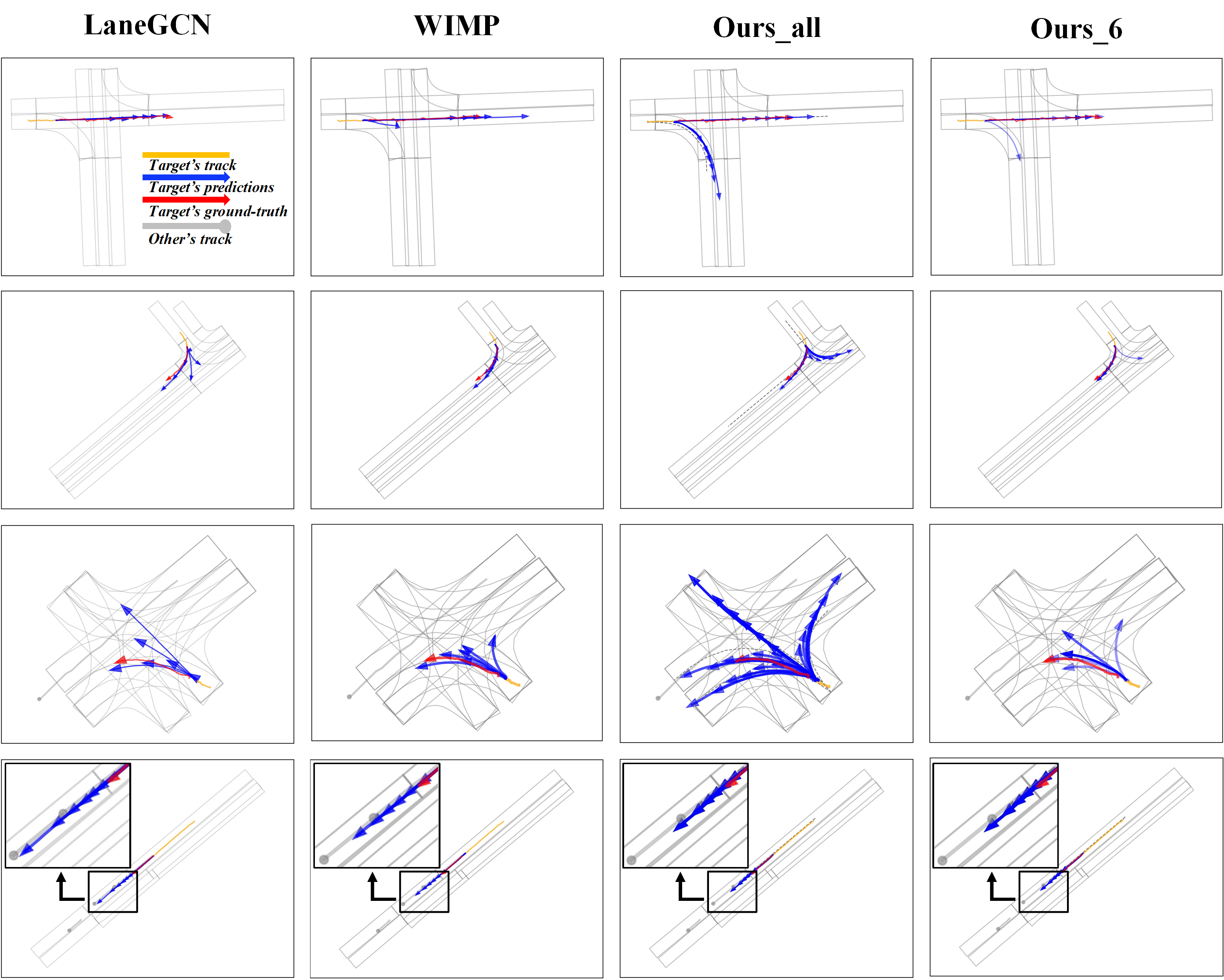}
\caption{Qualitative Results of the predicted trajectories obtained with Our model and baselines on Argoverse dataset. The road map and social agents’ history trajectories are depicted by light grey segments. The target agent’s history trajectory is shown in orange and ground-truth future trajectory in red. The future trajectory set produced by models are painted in blue. The results from left to right are LaneGCN, WIMP, ${\rm{Ours\_all}}$ and ${\rm{Ours\_6}}$. ${\rm{Ours\_all}}$ in the third column visualizes all the trajectories generated for all reachable lanes(dotted lines in figure). ${\rm{Ours\_6}}$ in right represents the remaining 6 typical trajectories filtered by selectors. From top to bottom, it contains difficult cases such as turns(line 1), forks(line 2), intersections(line 3), decelerate(line 4).}
\label{fig6}
\end{figure*}  
Fig.~\ref{fig6} shows the qualitative results of our model to other competitive methods on particular hard scenarios selected from Argoverse dataset. ${\rm{Ours\_all}}$ visualizes all trajectories generated by our model for all reachable lanes and ${\rm{Ours\_6}}$
shows the remaining trajectories filtered by two selectors. Our model could generate trajectories covering all plausible modes in each challenging scenario. The trajectories with reasonable probabilities produced by ${\rm{Ours\_6}}$ show the importance of two selectors to identify typical trajectories. 

Specifically from Fig.~\ref{fig6}, in turning and forking cases, it is hard to distinguish whether the agent will go on or turn to another path, the baselines only maintain current motion but our model can suggest both paths even there is only a small possibility. In intersection scenario, although there are more possible paths agent can choose, our model can still explicitly predict map-adaptive trajectories compared with others. Moreover, when there is congestion at the front, only our model correctly predicts the deceleration and tend to bypass. It reveals the effectiveness of jointly learning the social agents and lanes information to help prediction more reasonable and safe. 
\section{Conclusion}
In this paper, we introduced JAL-MTP, a novel mothod for multimodal trajectory prediction in complex driving scenarios. JAL-MTP captures the relation between the lanes and the motion of agents through the joint representation obtained for each instance-level lanes. Then JAL-MTP model the evolution of the current motion on these instance-level lanes to perform lane-based prediction, which help trajectories to be map-adaptive. In addition, we propose two self-supervised classification subtasks to guide our model to select the reasonable trajectories. The experiments conducted on Argoverse datasets demonstrate that the JAL-MTP could produce reasonable and accurate prediction in challenging scenarios, which is crucial for the safety and comfort of SDVs. In future work, the process to find approachable lanes by heuristic methods should be extended in neural network to achieve end-to-end solution without preprocessing.

\bibliographystyle{IEEEtran}
\bibliography{root} %change to your file name (with suffix .bib)

\end{document}